\pdfoutput=1

\documentclass[11pt]{article}

\usepackage{naacl2021}

\usepackage{times}
\usepackage{latexsym}

\usepackage[T1]{fontenc}

\usepackage[utf8]{inputenc}

\usepackage{microtype}

\usepackage{latexsym} 
\usepackage{graphicx}
\usepackage{wrapfig}
\usepackage{amssymb}
\usepackage{color,xcolor,colortbl}
\usepackage{enumitem}
\usepackage{soul}
\usepackage{amsmath}
\usepackage{url}
\usepackage{algorithm}
\usepackage[noend]{algpseudocode}
\usepackage{booktabs}
\usepackage{bm,bbm}
\usepackage{mathtools}
\usepackage{array}
\usepackage{multirow}
\usepackage{subcaption}
\usepackage{pbox}
\usepackage{pifont}
\usepackage{arydshln}
\usepackage{dblfloatfix}
\usepackage{relsize}
\usepackage{setspace}

\usepackage{pifont}
\newcommand{\cmark}{\ding{52}}%
\newcommand{\xmark}{\ding{56}}%
\newcommand{\done}{\rlap{$\square$}{\raisebox{0pt}{\large\hspace{-3.5pt} \textsf{x}}}%
\hspace{2pt}}

\newcommand\Bstrut{\rule[-0.9ex]{0pt}{0pt}}  

\newcommand\numberthis{\addtocounter{equation}{1}\tag{\theequation}}

\definecolor{darkblue}{rgb}{0.0, 0.0, 0.55}
\newenvironment{fontpbk}{\fontfamily{lmss}\selectfont}{\par} 
\setulcolor{blue} 
\usepackage[T1]{fontenc}
\usepackage[scaled=.8]{beramono}

\usepackage{soul}

\definecolor{d}{HTML}{c6dbef}
\definecolor{l}{HTML}{e5f5e0}
\definecolor{mygreen}{rgb}{0.13, 0.55, 0.13}

\title{A New Approach to Overgenerating and Scoring Abstractive Summaries}

\author{
Kaiqiang Song,$^1$
Bingqing Wang,$^2$
Zhe Feng,$^2$
Fei Liu$^1$\\[0.5em]
$^1$University of Central Florida, Orlando, FL\\
$^2$Robert Bosch LLC, Sunnyvale, CA\\[0.5em]
\texttt{kqsong@knights.ucf.edu, \{bingqing.wang,zhe.feng2\}@us.bosch.com}\\
\texttt{feiliu@cs.ucf.edu}\\
}

\begin{document}
\maketitle
\begin{abstract}

We propose a new approach to generate multiple variants of the target summary with diverse content and varying lengths, then score and select \emph{admissible} ones according to users' needs.
Abstractive summarizers trained on single reference summaries may struggle to produce outputs that achieve multiple desirable properties, i.e., capturing the most important information, being faithful to the original, grammatical and fluent.
In this paper, we propose a two-staged strategy to generate a diverse set of candidate summaries from the source text in stage one, then score and select admissible ones in stage two. 
Importantly, our generator gives a precise control over the length of the summary, which is especially well-suited when space is limited.
Our selectors are designed to predict the optimal summary length and put special emphasis on faithfulness to the original text.
Both stages can be effectively trained, optimized and evaluated.
Our experiments on benchmark summarization datasets suggest that this paradigm can achieve state-of-the-art performance.

\end{abstract}

\section{Introduction}
\label{sec:intro}

The learning objective of a modern abstractive summarizer is to produce system outputs that resemble reference summaries on a word-to-word basis.
It does \emph{not} promote outputs that possess multiple desirable properties, i.e., capturing the most important information, being faithful to the original text, grammatical and fluent, though some of these properties are exhibited by system abstracts as a natural outcome of a learned summarizer~\cite{see-etal-2017-get,takase-etal-2016-neural,tan-etal-2017-abstractive,chen-bansal-2018-fast,celikyilmaz-etal-2018-deep,gehrmann-etal-2018-bottom,liu-lapata-2019-hierarchical,lebanoff-etal-2019-scoring,fabbri-etal-2019-multi,brazinskas-etal-2020-shot}.
Without direct optimization of desired properties, system abstracts often change the meaning of the original document or fail to convey the main concepts~\cite{kryscinski-etal-2020-evaluating}.

\begin{table}
\centering
\setlength{\tabcolsep}{3pt}
\renewcommand{\arraystretch}{1.15}
\centering
\begin{fontpbk}
\begin{scriptsize}
\begin{tabular}[t]{|ll|}
\hline
\multicolumn{2}{|l|}{\textbf{Source Text}}\\
\hdashline
$\bullet$ & Police arrested five anti-nuclear protesters Thursday after they\\
& sought to disrupt loading of a French Antarctic research and\\
& supply vessel, a spokesman for the protesters said.\\
\hline
\hline
\multicolumn{2}{|l|}{\textbf{Summary}}\\
\hdashline
\textcolor{mygreen}{\cmark} & Police arrest anti-nuclear protesters\\
\textcolor{mygreen}{\cmark} & Protesters target French research ship\\
\textcolor{red}{\xmark} & French police arrest five anti-nuclear protesters\\
\textcolor{red}{\xmark} & Police arrest five anti-nuclear protesters in Antarctica\\
\textcolor{red}{\xmark} & Police arrest five anti-nuclear protesters at French Antarctic\\
\hline
\end{tabular}
\end{scriptsize}
\end{fontpbk}
\vspace{-0.05in}
\caption{
Example of alternative summaries generated from the source text.
\emph{Admissible} summaries are marked by \textcolor{mygreen}{\cmark}.
System summaries that fail to preserve the meaning of the source input are marked by \textcolor{red}{\xmark}.
}
\label{tab:summaries}
\vspace{-0.1in}
\end{table}

In this paper, we propose a new approach to over-generate and select \emph{admissible} summaries, 
which allows a summarizer to juggle multiple objectives and strike a good balance between them~\cite{belz-reiter-2006-comparing}.
Our approach consists of two stages.
Given a source text, a \emph{generator} explores the space of all possible lengths to produce multiple variants of the target summary that contain diverse content.
We then devise \emph{selectors} to validate the quality of alternative summaries to predict whether they are admissible.
Our selection mechanism can be customized to suit particular needs without changing the generation space.
Both stages can be effectively trained, optimized and evaluated.

\begin{table*}
\setlength{\tabcolsep}{2pt}
\renewcommand{\arraystretch}{1.15}
\centering
\begin{fontpbk}
\begin{scriptsize}
\begin{tabular}{|l||lllll|}
\hline
\multicolumn{6}{|l|}{\textbf{Source Text:}\quad A court here Thursday sentenced a 24-year-old man to 10 years in jail after he admitted pummelling his baby}\\
\multicolumn{6}{|l|}{\quad\quad\quad\quad\quad\quad\, son to death to silence him while watching television.}\\
\hline
\hline
\textbf{Left to Right Generation (\textcolor{red}{1 Summary})} & \multicolumn{5}{l|}{\textbf{Confidence Driven Generation (\textcolor{red}{4 Summaries})}}\\
Man \colorbox{yellow!30}{\underline{who}} & Man & & & & \colorbox{yellow!30}{\underline{gets 10 years}}\\
Man who \colorbox{yellow!30}{\underline{killed}} \quad [\dots] & Man & \colorbox{yellow!30}{\underline{who kill the baby}} & & & gets 10 years\\
Man who killed baby to hear television better gets \colorbox{yellow!30}{\underline{10}} & Man & who kill the baby & \colorbox{yellow!30}{\underline{to hear television}} & & gets 10 years\\
Man who killed baby to hear television better gets 10 \colorbox{yellow!30}{\underline{years}} & Man & who kill the baby & to hear television & \colorbox{yellow!30}{\underline{better}} & gets 10 years\\
\hline
\end{tabular}
\end{scriptsize}
\end{fontpbk}
\caption{
An example of the difference between left-to-right and confidence-driven summary generation. 
(\textsc{Left}) A single summary is produced in a left-to-right order.
(\textsc{Right}) Four summaries are generated in a confidence-driven mode. 
The most confident words are generated first, less vital ones later. 
Our generator learns to dynamically add or remove content given a target length to produce summaries of varying lengths—short, medium and long. 
The output is a diverse set of alternative summaries.
}
\label{tab:flex_gen}
\end{table*}

Crucially, we take a confidence-driven approach to summary generation rather than using a left-to-right order. 
Beginning writers and language learners do not write in a strict sequential manner. 
In a similar vein, our generator produces a summary by ``filling-in-the-blanks'' with appropriate words.
The most confident words are generated first, less vital ones later.
With confidence-driven generation, our summarizer learns to dynamically add or remove content, and even paraphrase to produce a summary of a given length.
In Table \ref{tab:flex_gen}, we show an example illustrating the difference between our method and left-to-right generation.
Our method dramatically enhances the capability of the generator, making it possible to explore summaries of varying lengths.

Identifying admissible summaries with desired properties is critical for a summarizer.
Summaries of very short lengths may fail to capture the main concepts, and this kind of incomplete or partial information can lead to false assumptions about the original content.
Moreover, summaries of moderate lengths may still contain hallucinated content that is nonexistent in the source text~\cite{maynez-etal-2020-faithfulness}.
We present two summary selectors to combat these issues.
Our first selector aims to predict what summary length is most suitable for a source text,
whereas a second selector puts special emphasis on the overall quality of the system summary, in particular its faithfulness to the original text~\cite{falke-etal-2019-ranking,durmus-etal-2020-feqa}.

A novel dataset has been introduced in this work where we associate a source text with multiple summaries, and \emph{admissible} ones are manually labelled by human annotators.
Not only can the dataset be used to judge the effectiveness of summary selectors, 
but it provides a new testbed for future summarizers to compare their outputs against multiple reference summaries, which is key to improve the reliability of evaluation results~\cite{louis-nenkova-2013-automatically}.
We have focused on generating abstractive summaries from single source sentences, but the insights gained from this study could inform the design of summarizers of all forms.
Our method also has a great potential to incorporate human-in-the-loop to teach the model to select the best summary.
The main contributions of this paper are:
\begin{itemize}[topsep=3pt,itemsep=-1pt,leftmargin=*]

    \item We propose a new approach to generate multiple variants of the target summary that have varying lengths, then score and select the best summaries according to our needs.
    
    \item Our generator controls over the length of the summary, which is especially well-suited when space is limited. Our selectors are designed to predict the optimal summary length and put special emphasis on faithfulness to the original text.
    
    \item Our experiments on benchmark summarization datasets suggest that this paradigm can surpass results of previous studies or rival state-of-the-art. We conclude with a discussion of our key findings, which has implications for the development of robust abstractive summarizers.\footnote{Our code and annotated data are made available on Github at \url{https://github.com/ucfnlp/varying-length-summ}}
    
\end{itemize}

\section{Related Work}
\label{sec:related}

It is important for neural abstractive summarizers to produce summaries that are faithful to the original texts~\cite{cao-etal-2017-fsum,kryscinski-etal-2019-neural,lebanoff-etal-2019-analyzing,wang-etal-2020-asking,dong-etal-2020-multi,zhang-etal-2020-optimizing}.
However, it remains questionable as to whether a summarizer must acquire that ability by learning from human reference summaries, or possibly through external resources such as textual entailment predictions~\cite{falke-etal-2019-ranking}.
In this paper, we present a two-stage strategy to over-generate, then score system summaries externally for faithfulness and overall quality.

\begin{figure*}[t]
\centering
\includegraphics[width=6in]{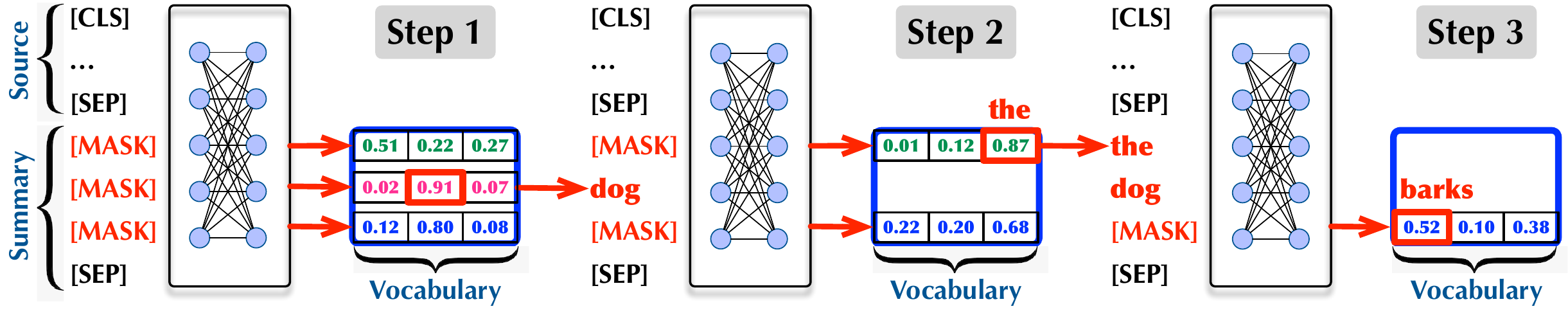}
\caption{
An illustration of the generation process.
A sequence of placeholders (``\textsc{[mask]}'') are placed following the source text.
Our model simultaneously predicts the most probable tokens for \emph{all positions}, rather than predicting only the most probable \emph{next} token in an autoregressive setting.
We obtain the token that has the highest probability, and use it to replace the \textsc{[mask]} token of that position.
Next, the model makes new predictions for all remaining positions, conditioned on the source text and \emph{all summary tokens seen thus far}.
Our generator produces a summary having the \emph{exact given length} and with a proper endpoint.
}
\label{fig:architecture}
\end{figure*}

Previous work has sought to control various aspects of the generated summary, including the style, length and amount of reused text~\cite{kikuchi-etal-2016-controlling,hu2017,fan-etal-2018-controllable,keskarCTRL2019,makino-etal-2019-global,Song:2020:Copy}.
In contrast, our generator focuses on producing \emph{multiple} variants of the target summary that have diverse content and varying lengths.
It offers precise control over the length of the summary, which has an important implication for fair comparison between different summarization systems~\cite{napoles-etal-2011-evaluating,shapira-etal-2018-evaluating}.

Our methodology allows for greater flexibility in designing summary selectors.
The selectors may allow multiple admissible summaries to be identified for any source input according to users' needs.
On the contrary, post-editing of system summaries through a set of basic operations such as insertion and deletion~\cite{NIPS2019_9297,malmi-etal-2019-encode,dong-etal-2019-editnts,correia-martins-2019-simple} may have intrinsic limitations by learning from single reference summaries to produce single outputs.
In this paper, we provide a new dataset where each source text is associated with multiple admissible summaries to encourage diverse outputs.

Our generator is inspired by unsupervised pretraining of deep neural models~\cite{peters-etal-2018-deep,radford2019language,devlin-etal-2019-bert,yan2020prophetnet,zhang2020pegasus,lewis-etal-2020-bart} and non-autoregressive machine translation~\cite{gu2018nonautoregressive,ghazvininejad-etal-2019-mask}.
Distinct from these is our confidence-driven generation that goes beyond left-to-right order.
It uses a denoising objective during training and is conveniently transformed into a semi-autoregressive generator at test time.
We introduce a customized beam search algorithm to promote the generation of diverse outputs.
In the following section, we describe in detail our two-step strategy.

\section{A Confidence-Driven Generator}
\label{sec:method}

We seek to produce a highly diverse set of alternative summaries from any source input, but standard neural language generators with beam search only produce high-likelihood sequences rather than diverse ones~\cite{ippolito-etal-2019-comparison}.
To address this limitation, we devise a new generator that is capable of producing summaries of varying lengths. 
A long summary can cover more important information of the source text, whereas a short summary is easy-to-read.
Moreover, it produces a summary having the \emph{exact given length} and with a proper endpoint.
This is achieved by shifting away from left-to-right generation but building a summary using a confidence-driven approach.

Our generator is illustrated in Figure~\ref{fig:architecture}.
To generate a summary of $L$ tokens, we place a number of \textsc{[mask]} tokens following the source text, which serve as ``placeholders'' for summary tokens.
Importantly, our generator simultaneously predicts the most probable tokens for \emph{all positions}, as opposed to predicting only the most probable \emph{next} token in an autoregressive setting.
We obtain the token that has the highest probability across all positions, and use it to replace the \textsc{[mask]} token of that position.
Next, the model continues to make predictions for all remaining positions, conditioned on the source text and the summary tokens seen thus far of varying positions.

Let $\mathbf{x}=\{x_i\}_{i=1}^N$ be the source and $\mathbf{y}=\{y_j\}_{j=1}^M$ the summary sequence.
Our confidence-driven generation process defines a new order of summary tokens, $\mathbf{o}=\{o_j\}_{j=1}^M$, $o_j \in [M]$, 
according to which $P_\theta(\mathbf{y}|\mathbf{x})$ is factorized into a product of conditional probabilities $P_\theta(y_{\textcolor{black}{o_j}}|y_{\textcolor{black}{\mathbf{o}_{<j}}},\mathbf{x})$ (Eq.~(\ref{eq:prob})),
where $\theta$ are model parameters to be optimized during training.
Our learning objective is to minimize the negative data log-likelihood (Eq.~(\ref{eq:loss})) to predict missing tokens $y_{\textcolor{black}{o_j}}^*$ conditioned on the source text $\mathbf{x}$ and the summary tokens seen thus far $y_{\textcolor{black}{\mathbf{o}_{<j}}}$.
\begin{align*}
\textstyle
P_\theta(\mathbf{y}|\mathbf{x};\mathbf{o}) &= \prod_{j=1}^M P_\theta(y_{\textcolor{black}{o_j}}|y_{\textcolor{black}{\mathbf{o}_{<j}}},\mathbf{x})
\numberthis\label{eq:prob}\\
\mathcal{L}(\theta) &= - \sum_{j=1}^M \log P_\theta(y_{\textcolor{black}{o_j}}^*|y_{\textcolor{black}{\mathbf{o}_{<j}}},\mathbf{x}) \numberthis\label{eq:loss}
\end{align*}

Our generator is trained with a denoising objective.
It consists of a decoder-only architecture with 12 Transformer blocks~\cite{NIPS2019_9464}.
Given a source text and a summary, we replace a portion of their tokens by the \textsc{[mask]} token, and the model is trained to reconstruct the original data from the corrupted text.
It differs from autoregressive models in that the context of each position can consist of tokens from both left and right—a source word can attend to other source words and a summary word can attend to source words and summary words seen thus far \emph{of varying positions}—hence capturing a bidirectional context.
The training procedure is thus analogous to that of permutation-based language modeling~\cite{NIPS2019_8812}.

\begin{table}[t]
\centering
\setlength{\tabcolsep}{3pt}
\renewcommand{\arraystretch}{1.2}
\centering
\begin{fontpbk}
\begin{scriptsize}
\begin{tabular}{l|l}
Input & \textbf{The Bank of Japan appealed to financial}\\
& \textbf{markets to remain calm Friday following the US decision}\\
& \textbf{to order Daiwa Bank Ltd. to close its US operations.}\\
\hline
\hline
\Bstrut L=6 & $\underbracket{\mbox{\textcolor{blue}{BoJ}}}_{2,5} \underbracket{\mbox{\textcolor{magenta}{calls}}}_{4} \underbracket{\mbox{\textcolor{magenta}{for}}}_{6} \underbracket{\mbox{\textcolor{magenta}{calm.}}}_{3,1}$\\ 
L=7 & $\underbracket{\mbox{\textcolor{blue}{BoJ}}}_{3,7} \underbracket{\mbox{\textcolor{magenta}{calls}}}_{4} \underbracket{\mbox{\textcolor{magenta}{for}}}_{5} \underbracket{\mbox{market}}_{6} \underbracket{\mbox{\textcolor{magenta}{calm.}}}_{2,1}$\\
L=8 & $\underbracket{\mbox{\textcolor{blue}{BoJ}}}_{5,7} \underbracket{\mbox{\textcolor{red}{urges}}}_{6} \underbracket{\mbox{\textcolor{red}{markets}}}_{4} \underbracket{\mbox{\textcolor{red}{to}}}_{3} \underbracket{\mbox{\textcolor{red}{remain}}}_{1} \underbracket{\mbox{\textcolor{red}{calm.}}}_{8,2}$\\
L=9 & $\underbracket{\mbox{\textcolor{blue}{BoJ}}}_{6,2} \underbracket{\mbox{\textcolor{red}{urges}}}_{7} \underbracket{\mbox{financial}}_{4} \underbracket{\mbox{\textcolor{red}{markets}}}_{5} \underbracket{\mbox{\textcolor{red}{to}}}_{9} \underbracket{\mbox{\textcolor{red}{remain}}}_{1} \underbracket{\mbox{\textcolor{red}{calm.}}}_{8,3}$\\
L=10 & $\underbracket{\mbox{\textcolor{blue}{BoJ}}}_{1,2} \underbracket{\mbox{\textcolor{magenta}{calls}}}_{6} \underbracket{\mbox{\textcolor{magenta}{for}}}_{7} \underbracket{\mbox{\textcolor{magenta}{calm}}}_{5} \underbracket{\mbox{\textcolor{cyan}{after}}}_{8} \underbracket{\mbox{\textcolor{cyan}{Daiwa}}}_{10,4} \underbracket{\mbox{\textcolor{cyan}{closure.}}}_{9,3}$\\
L=11 & $\underbracket{\mbox{\textcolor{blue}{BoJ}}}_{1,2} \underbracket{\mbox{\textcolor{magenta}{calls}}}_{6} \underbracket{\mbox{\textcolor{magenta}{for}}}_{7} \underbracket{\mbox{\textcolor{magenta}{calm}}}_{5} \underbracket{\mbox{\textcolor{cyan}{after}}}_{8} \underbracket{\mbox{\textcolor{cyan}{Daiwa}}}_{11,4} \underbracket{\mbox{\textcolor{cyan}{Bank}}}_{3} \underbracket{\mbox{\textcolor{cyan}{closure.}}}_{10,9}$\\
L=12 & $\underbracket{\mbox{\textcolor{blue}{BoJ}}}_{2,3} \underbracket{\mbox{\textcolor{magenta}{calls}}}_{5} \underbracket{\mbox{\textcolor{magenta}{for}}}_{6} \underbracket{\mbox{\textcolor{magenta}{calm}}}_{1} \underbracket{\mbox{\textcolor{cyan}{after}}}_{11} \underbracket{\mbox{\textcolor{cyan}{Daiwa}}}_{8,7} \underbracket{\mbox{\textcolor{cyan}{Bank}}}_{9} \underbracket{\mbox{\textcolor{cyan}{closure}}}_{12} \underbracket{\mbox{order.}}_{10,4}$\\
L=13 & $\underbracket{\mbox{\textcolor{blue}{BoJ}}}_{6,13} \underbracket{\mbox{\textcolor{red}{urges}}}_{8} \underbracket{\mbox{\textcolor{red}{markets}}}_{7} \underbracket{\mbox{\textcolor{red}{to}}}_{9} \underbracket{\mbox{\textcolor{red}{remain}}}_{11} \underbracket{\mbox{\textcolor{red}{calm}}}_{4} \underbracket{\mbox{\textcolor{cyan}{after}}}_{10} \underbracket{\mbox{\textcolor{cyan}{Daiwa}}}_{5,2} \underbracket{\mbox{\textcolor{cyan}{Bank}}}_{1} \underbracket{\mbox{\textcolor{cyan}{closure.}}}_{12,3}$\\
L=14 & $\underbracket{\mbox{\textcolor{blue}{BoJ}}}_{3,4} \underbracket{\mbox{\textcolor{magenta}{calls}}}_{7} \underbracket{\mbox{\textcolor{magenta}{for}}}_{8} \underbracket{\mbox{\textcolor{magenta}{calm}}}_{2} \underbracket{\mbox{\textcolor{cyan}{after}}}_{14} \underbracket{\mbox{\textcolor{cyan}{Daiwa}}}_{13,6} \underbracket{\mbox{\textcolor{cyan}{Bank}}}_{5} \underbracket{\mbox{'s}}_{10,9} \underbracket{\mbox{US}}_{11} \underbracket{\mbox{\textcolor{cyan}{closure.}}}_{12,1}$\\
L=15 & $\underbracket{\mbox{\textcolor{blue}{BoJ}}}_{10,3} \underbracket{\mbox{\textcolor{magenta}{calls}}}_{4} \underbracket{\mbox{\textcolor{magenta}{for}}}_{5} \underbracket{\mbox{\textcolor{magenta}{calm}}}_{2} \underbracket{\mbox{\textcolor{orange}{after}}}_{15} \underbracket{\mbox{\textcolor{orange}{US}}}_{8} \underbracket{\mbox{\textcolor{orange}{order}}}_{13} \underbracket{\mbox{for}}_{14} \underbracket{\mbox{Daiwa}}_{9,6} \underbracket{\mbox{Bank}}_{7} \underbracket{\mbox{to}}_{11} \underbracket{\mbox{close.}}_{12,1}$\\
L=16 & $\underbracket{\mbox{\textcolor{blue}{BoJ}}}_{3,5} \underbracket{\mbox{\textcolor{magenta}{calls}}}_{4} \underbracket{\mbox{\textcolor{magenta}{for}}}_{7} \underbracket{\mbox{\textcolor{magenta}{calm}}}_{2} \underbracket{\mbox{\textcolor{orange}{after}}}_{16} \underbracket{\mbox{\textcolor{orange}{US}}}_{13} \underbracket{\mbox{\textcolor{orange}{order}}}_{12} \underbracket{\mbox{on}}_{14} \underbracket{\mbox{Daiwa}}_{8,6} \underbracket{\mbox{'s}}_{11,10} \underbracket{\mbox{US}}_{9} \underbracket{\mbox{operations.}}_{15,1}$\\
\end{tabular}
\end{scriptsize}
\end{fontpbk}
\caption{
The target summary length $\textsf{L}$ is adjusted to produce alternative summaries that have diverse content.
Our generator can dynamically add or remove content, and paraphrase to produce a summary of a given length.
The numbers indicate the order in which the summary tokens are generated. 
``BoJ'' stands for ``Bank of Japan''.
It maps to two tokens according to Byte Pair Encoding (BPE).
Each summary has an ending period, so the last word also maps to two tokens.
}
\label{tab:example_order}
\end{table}

Our training schedule begins with masking out 10\% of source tokens and linearly decreases it to 0\% throughout training steps.
Masking out a portion of source tokens helps the model learn contextualized representations given bidirectional context.
On the target side, the schedule begins with masking out 90\% of summary tokens and linearly decreases it to 60\%. 
It allows the model to learn to predict missing summary tokens and copy source tokens to the summary.
When a token is chosen, it is replaced with the \textsc{[mask]} token 80\% of the time, a random token of the vocabulary 10\% of the time, and remains unchanged otherwise.

In Table~\ref{tab:example_order}, we present example summaries produced by our new confidence-driven generator for a source input.
The summaries have varying lengths and degrees of details.
Our generator learns to add or remove content, and even paraphrase to produce a summary of a given length.
We adjust the target summary length ($L$) to produce diverse summaries.
Moreover, there exists more than one admissible summaries that capture the important information of the source text, while being grammatical and faithful to the original. 
It is important to note that, to decode the best summary of length $L$, our generator requires a \emph{position-aware} beam search algorithm to explore the space of candidate summaries, which is described next.

\subsection{Position-Aware Beam Search}
\label{sec:decode}

\begin{figure}
\centering
\begin{minipage}[h]{0.48\textwidth}
\begin{algorithm}[H]
\caption{Position-Aware Beam Search}\label{alg:Beam}
\begin{algorithmic}[1]
\smaller
\Procedure{PosAwareBeam}{$\mbox{SourceText}$, $L$, $K$}\\
\Comment{\textcolor{darkblue}{$L$ is the summary length and $K$ is the beam size.}}
\State $\mathcal{S}_0 \gets \{\mbox{\textsc{[Mask]}}\times L\}$ \quad\Comment{\textcolor{darkblue}{Initial summary.}}
\State $\mathcal{M}_0 \gets [\mathbf{1}]_{L \times |\mathcal{V}|}$ \,\,\,\Comment{\textcolor{darkblue}{A binary mask of $L$ positions.}}
\State $\mathcal{H} \gets \{(0, \mathcal{S}_0, \mathcal{M}_0)\}$ \,\,\;\Comment{\textcolor{darkblue}{A priority queue.}}
\For{$j = 1, \dots, L$}
    \State $\textit{Candidates} \gets \{\}$
    \For{$\textit{hyp} \in \mathcal{H}$}
        \State $\textit{score}', \mathcal{S}', \mathcal{M}' \gets \textit{hyp}$\\
        \quad\quad\quad\quad\,\,\Comment{\textcolor{darkblue}{Estimate token probabilities.}}
        \State $\mathcal{P}_{L \times |\mathcal{V}|} \gets \mbox{Gen}(\mbox{SourceText}, \mathcal{S}')$ 
        \State $\mathcal{P}' \gets \mathcal{P} \odot \mathcal{M}'$\\ 
        \quad\quad\quad\quad\,\,\Comment{\textcolor{darkblue}{Record $K$-best tokens and positions.}}
        \For {$s_k, w_k, p_k \in \textit{Top-K-Scores}(\mathcal{P}')$}
            \State $\textit{score}'' \gets \textit{score}' + s_k$
            \State $\mathcal{S}'' \gets \textit{replace}(\mathcal{S}', p_k, w_k)$
            \State $\mathcal{M}'' \gets \textit{replace}(\mathcal{M}', p_k, [\mathbf{0}]_{\mathbf{1}\times|\mathcal{V}|})$
            \State $\textit{Candidates}.add((\textit{score}'', \mathcal{S}'', \mathcal{M}''))$
        \EndFor
    \EndFor
    \State $\mathcal{H} \gets \textit{Top-K-Scores}(\textit{Candidates})$
\EndFor\\
\Return $\mathcal{H}_0$ \,\,\Comment{\textcolor{darkblue}{The best summary of length $L$.}}
\EndProcedure
\end{algorithmic}
\label{alg:beam_search}
\end{algorithm}
\end{minipage}
\end{figure}

\begin{table*}
\setlength{\tabcolsep}{3pt}
\renewcommand{\arraystretch}{1.2}
\begin{minipage}[t]{0.575\hsize}
\centering
\begin{fontpbk}
\begin{scriptsize}
\begin{tabular}[t]{|lp{3.2in}|}
\hline
\multicolumn{2}{|l|}{\textbf{Entity Replacement}}\\
\hdashline
$\bullet$ & German art experts have authenticated a painting believed to be the last portrait ever made of the composer Wolfgang Amadeus Mozart, the body which runs Berlin's museums said on Thursday.\\
\hdashline
\textcolor{mygreen}{\cmark} & German experts identify last known portrait of \textcolor{red}{\textbf{Mozart}}\\
\textcolor{red}{\xmark} & German experts identify last known portrait of \textcolor{red}{\textbf{Mount Mayon's}}\\
\hline
\hline
\multicolumn{2}{|l|}{\textbf{Negation}}\\
\hdashline
$\bullet$ & US Secretary of State Condoleezza Rice suggested Tuesday that International Atomic Energy Agency chief Mohamed ElBaradei should not interfere in diplomatic issues after he warned against the hasty use of force in the Iranian nuclear dispute.\\
\hdashline
\textcolor{mygreen}{\cmark} & Rice suggests IAEA chief \textcolor{red}{\textbf{should}} stay clear of diplomacy\\
\textcolor{red}{\xmark} & Rice suggests IAEA chief \textcolor{red}{\textbf{shouldn't}} stay clear of diplomacy\\
\hline
\hline
\multicolumn{2}{|l|}{\textbf{Incomplete Summary}}\\
\hdashline
$\bullet$ & Total Hong Kong dollar deposits grew 2.2 percent in March, compared to 2.1 percent in February, according to the Hong Kong Monetary Authority.\\
\hdashline
\textcolor{mygreen}{\cmark} & HK Bank Deposits \textcolor{red}{\textbf{Increase in March}}\\
\textcolor{red}{\xmark} & \textcolor{red}{\textbf{Increase in March}}\\
\hline
\end{tabular}
\end{scriptsize}
\end{fontpbk}
\end{minipage}
\hfill
\begin{minipage}[t]{0.415\hsize}
\centering
\begin{fontpbk}
\begin{scriptsize}
\begin{tabular}[t]{|lp{2.3in}|}
\hline
\multicolumn{2}{|l|}{\textbf{Search and Replace}}\\
\hdashline
$\bullet$ & Israel is on course to complete the main tranche of its controversial West Bank security barrier in 2004 and wrap up the project in the following year, the defence ministry said Wednesday\\
\hdashline
\textcolor{mygreen}{\cmark} & \textcolor{mygreen}{\textbf{Israel surges ahead with West Bank barrier construction}}\\
\textcolor{red}{\xmark} & \textcolor{blue}{\textbf{Soul-searching in Israel over shooting of West Bank barrier protestor}}\\
\hline
\hline
\multicolumn{2}{|l|}{\textbf{Swap Segments}}\\
\hdashline
$\bullet$ & The Security Council on Thursday voted unanimously to extend the mandate of the UN mission in Georgia for four months ahead of next week's international talks on the fallout of the recent Caucasus conflict.\\
\hdashline
\textcolor{mygreen}{\cmark} & \textcolor{mygreen}{\textbf{Security Council extends mandate of}} \textcolor{blue}{\textbf{UN mission in Georgia}}\\
\textcolor{red}{\xmark} & \textcolor{blue}{\textbf{UN mission in Georgia}} \textcolor{mygreen}{\textbf{Security Council extends mandate of}}\\
\hline
\end{tabular}
\end{scriptsize}
\end{fontpbk}
\end{minipage}
\caption{
Corruption types. 
A positive instance for the selector consists of a ground-truth summary (marked by \textcolor{mygreen}{\cmark}) and its source text.
A negative instance consists of a corrupted summary (\textcolor{red}{\xmark}) and its source text. 
\emph{Entity Replacement}: replacing a named entity of the ground-truth summary with a random entity.
\emph{Negation}: negating a ground-truth summary sentence.
\emph{Incomplete Summary}: replacing the ground-truth summary with one of its sentence constituents to produce a corrupted summary that contains 5 words or less. 
\emph{Search and Replace}: swapping the ground-truth summary with a similar summary in the training set that have 4 or more common bigrams.
\emph{Swap Segments}: splitting the ground-truth into two parts of similar length, the parts are swapped to produce an ungrammatical summary.
}
\label{tab:corrupted}
\end{table*}

A position-aware beam of size $K$ not only contains the $K$-best candidate summaries having the highest log-likelihood at any time step, but it also records the positions of summary tokens seen thus far for each candidate summary.
The tokens of candidate summaries can be decoded in any order and occur in different positions, marking an important distinction between position-aware and traditional beam search~\cite{meister-etal-2020-beam}.
The method is realized by associating each candidate summary with a binary matrix $\mathcal{M} \in \{0,1\}_{L \times |\mathcal{V}|}$, which records what positions have been filled by which summary tokens and what positions remain available.

Concretely, we use $\mathcal{S}'$ to denote a candidate summary, $score'$ is its data log-likelihood and $\mathcal{M}'$ is a binary mask (Line 9).
Our generator predicts the token probabilities $\mathcal{P}_{L\times|\mathcal{V}|}$ for all positions, conditioned on the source text and the summary tokens seen thus far.
The binary mask $\mathcal{M}'$ indicates positions that remain available (Line 11--12).
We obtain the top-$K$ tokens that have the highest probability scores across all positions, record their summary hypotheses and likelihood scores.
These positions are then marked as taken (Line 14--18).

The decoding process continues until all of the $L$ positions are filled by summary tokens.
This makes our method different from traditional beam search, the latter terminates when an end-of-sequence symbol \textsc{[Sep]} is generated for the summary.
Particularly, our method is advantageous as it exerts \emph{precise control} over the summary length. 
The model learns to decide what content to be included in the summary given the limited space available, yielding summaries with varying degrees of details.

\section{The Selectors}
\label{sec:selector}

We present two selectors to respectively assess the overall quality of the summary and predict the optimal summary length.
Our selectors assume the role of a responsible agent that, when provided with a source text and multiple alternative summaries, can effectively recognize the \emph{admissible} ones.
It has the potential to incorporate human-in-the-loop in future to teach the model to select best summaries.

\subsection{Best Overall Quality}

Our goal is to build a selector to discern the difference between high and low-quality summaries.
In an ideal scenario, we have human annotators to vet each source text/summary pair, the annotated data are used to train the selector. The process, however, is both expensive and time-consuming. 
Inspired by Kryściński et al.~\shortcite{kryscinski-etal-2020-evaluating}, we automatically construct a large number of minimally different pairs, where a positive instance comprises of the source text and its ground-truth summary, and a negative instance includes the source text and a corrupted summary.
We experiment with various means to generate corrupted summaries from a ground-truth summary. 
The corruptions should resemble common mistakes made by neural abstractive summarizers, including generating factually incorrect details, failing to convey the main points of the source text, and being ungrammatical. 
The corruption types experimented in this paper are illustrated in Table~\ref{tab:corrupted}.

Distinguishing our work from that of Kryściński et al.~\shortcite{kryscinski-etal-2020-evaluating} are
(i) \emph{Search and Replace}, we swap the ground-truth summary with a similar summary in the training set that have $\ge$4 common bigrams to form a negative instance.
(ii) \emph{Swap Segments} splits a ground-truth summary into two parts of similar lengths, then swaps them to produce an ungrammatical summary.
(iii) \emph{Incomplete Summary} replaces a ground-truth summary by one of its sentence constituents, yielding a corrupted summary that fails to convey the main ideas.
These corruptions are designed to emulate system summaries that are too short to capture the main concepts, or contain hallucinated content that is not found in the source text.

We next build a binary classifier to predict if a summary is admissible given the source text.
To distill information from the source text and the summary, we encode them into hidden vectors using RoBERTa~\cite{liu2019roberta}.
These are denoted by $\mathbf{h}_\mathbf{x}$ and $\mathbf{h}_\mathbf{y}$, respectively.
We create a vector for the pair,
$\mathbf{h} = \mathbf{h}_\mathbf{x} \oplus \mathbf{h}_\mathbf{y} \oplus (\mathbf{h}_\mathbf{x} - \mathbf{h}_\mathbf{y}) \oplus (\mathbf{h}_\mathbf{x} * \mathbf{h}_\mathbf{y})$, 
consisting of a concatenation of the two hidden vectors, their absolute difference $(\mathbf{h}_\mathbf{x} - \mathbf{h}_\mathbf{y})$ and their element-wise product $(\mathbf{h}_\mathbf{x} * \mathbf{h}_\mathbf{y})$.
$\oplus$ is a concatenation of vectors.
The output vector $\mathbf{h}$ is expected to capture the gist of the source text and the summary,
and a similar approach is being used for natural language inference~\cite{chen-etal-2018-neural}.
The vector $\mathbf{h}$ is fed to a feed-forward layer to predict whether the summary is admissible given the source text.
We have chosen to design the selector as a classifier rather than a ranking model because there can exist multiple, equally valid summaries for any source input.
The classifier allows us to identify admissible summaries that are not only true-to-original but has the best overall quality.

\subsection{Best Summary Length}

Finding a suitable length for the summary is one of the most important open problems in automatic summarization~\cite{shapira-etal-2018-evaluating,sun-etal-2019-compare}.
A summary should be shorter than the original, but long enough to include the most important information.
Length normalization seeks to rescale the log-likelihood score of a summary, denoted by $\mathcal{S}(\mathbf{x}, \mathbf{y}) = \log p_\theta(\mathbf{y}|\mathbf{x})$, by its length $|\mathbf{y}|$, with an exponent $p$ (Eq.~(\ref{eq:ln})).
It is used by some neural abstractive summarizers~\cite{see-etal-2017-get,lewis-etal-2020-bart}.
However, the method does not consider the density of information in the source text and it may still generate ultra-short summaries. 
\begin{align*}
\mathcal{S}_{ln}(\mathbf{x},\mathbf{y}) &= \mathcal{S}(\mathbf{x},\mathbf{y})/|\mathbf{y}|^p
\numberthis\label{eq:ln}
\end{align*}

Instead, we attempt to estimate the appropriate length of the summary given a source text, denoted by $\mathcal{L}_{\mbox{\tiny pred}}$, and reward a system summary if it stays close to the estimated length~\cite{huang-etal-2017-finish}.
Concretely, we assign a per-word reward to the summary, represented by $r \min(|\mathbf{y}|, \mathcal{L}_{\mbox{\tiny pred}})$ (Eq.~(\ref{eq:length-bounded-reward})).
A system summary continues to be rewarded until it reaches the predicted length ($|\mathbf{y}| \le \mathcal{L}_{\mbox{\tiny pred}}$).
Beyond that, increasing the length of the summary does not lead to additional rewards. 
We obtain the predicted length $\mathcal{L}_{\mbox{\tiny pred}}$ using a baseline abstractive summarizer, which takes the source text as input and greedily decodes a summary in a left-to-right manner until an end-of-sequence symbol is predicted; $\mathcal{L}_{\mbox{\tiny pred}}$ is the length of the decoding sequence.
$r$ is a coefficient to scale the reward and it is tuned on the validation data.
Finally, the reward-augmented log-likelihood $\mathcal{S}_{rwd}(\mathbf{x},\mathbf{y})$ is used as a scoring function to rank all summary hypotheses of varying lengths.
\begin{align*}
\mathcal{S}_{rwd}(\mathbf{x},\mathbf{y}) = \mathcal{S}(\mathbf{x},\mathbf{y}) + r \min(|\mathbf{y}|, \mathcal{L}_{\mbox{\tiny pred}})
\numberthis\label{eq:length-bounded-reward}
\end{align*}

\section{Experiments}

\noindent\textbf{Datasets}\quad
We perform extensive experiments on Gigaword~\cite{Parker:2011} and Newsroom~\cite{grusky-etal-2018-newsroom} datasets.
The goal is to generate an abstractive summary from a lengthy source sentence.
For each article, we pair its first sentence with the title to form a summarization instance.
Both datasets contain large collections of news articles.
Gigaword (1995--2010) contains 3,810,674 / 10,000 / 1,951 instances, respectively, in the train, validation and test splits. Newsroom (1998--2017) contains 199,341 / 21,530 / 21,377 instances, respectively.
We conduct experiments on both datasets to demonstrate the generality of our two-staged strategy.
Our method generates a diverse set of summaries from a source sentence in stage one, then score and select admissible summaries in stage two.

\begin{table}[t]
\centering
\setlength{\tabcolsep}{5pt}
\renewcommand{\arraystretch}{1.2}
\begin{fontpbk}
\begin{scriptsize}
\begin{tabular}{|l|ccc|}
\hline
\textbf{System} & \textbf{R-1} & \textbf{R-2} & \textbf{R-L}\\
\hdashline
lvt2k-1sent {\scriptsize\cite{nallapati-etal-2016}} & 32.67 & 15.59 & 30.64 \\
SEASS {\scriptsize\cite{zhou-etal-2017-selective}} & 36.15 & 17.54 & 33.63 \\
DRGD {\scriptsize\cite{li-etal-2017-deep}} & 36.27 & 17.57 & 33.62 \\
Pointer-Gen {\scriptsize\cite{see-etal-2017-get}} & 34.19 & 16.92 & 31.81\\
R3Sum {\scriptsize\cite{cao-etal-2018-retrieve}} & 37.04 & 19.03 & 34.46 \\
EntailGen {\scriptsize\cite{guo-etal-2018-soft}} & 35.98 & 17.76 & 33.63\\
BiSET {\scriptsize\cite{wang-etal-2019-biset}} & 38.45 & 19.53 & 36.04 \\
MASS {\scriptsize\cite{song2019mass}} & 38.73 & 19.71 & 35.96\\
UniLM {\scriptsize\cite{NIPS2019_9464}} & 38.90 & 20.05 & 36.00\\
PEGASUS {\scriptsize\cite{zhang2020pegasus}} & 39.12 & 19.86 & 36.24\\
\hdashline
Ours (\textsc{Average}) & 35.51 & 16.33 & 32.75\\
Ours (\textsc{Best Quality}) & 36.71 & 17.27 & 33.63\\
Ours (\textsc{Best Summary Length}) & \textbf{39.27} & \textbf{20.40} & \textbf{36.76}\\
\hline
\end{tabular}
\end{scriptsize}
\end{fontpbk}
\caption[Caption for LOF]{
Results on the Gigaword test set evaluated by ROUGE~\cite{lin-2004-rouge}.\protect\footnotemark
}
\label{tab:results_gigaword}
\vspace{-0.1in}
\end{table}
\footnotetext{
Our experiments are performed on the original Gigaword dataset~\cite{Parker:2011} without anonymization.
The data provided by Rush et al.~\shortcite{rush-etal-2015-neural} replaced all digit characters with \# and replaced word types seen less than 5 times with UNK.
}

\begin{table*}[t]
\setlength{\tabcolsep}{6pt}
\renewcommand{\arraystretch}{1.2}
\centering
\begin{fontpbk}
\begin{scriptsize}
\begin{tabular}{|l|ccccccccccc|c|c|}
\hline
& \multicolumn{11}{c|}{\textbf{Summary Length (L)}} & \textbf{Best} & \textbf{Best} \\
\textbf{Gigaword} & 7 & 8 & 9 & 10 & 11 & 12 & 13 & 14 & 15 & 16 & \textbf{Avg.} & \textbf{Quality} & \textbf{Length}\\
\hdashline
\textbf{R-1 F$_1$} (\%) & 32.01 & 35.42 & 37.05 & 37.95 & \textbf{38.05} & 37.79 & 37.27 & 36.66 & 35.75 & 35.13 & 36.31 & 36.71 & \textbf{39.27}\\
\textbf{R-2 F$_1$} (\%) & 13.47 & 15.68 & 17.39 & 18.31 & \textbf{18.24} & 18.22 & 17.85 & 17.19 & 16.63 & 16.00 & 16.90 & 17.27 & \textbf{20.40}\\
\textbf{R-L F$_1$} (\%) & 29.76 & 32.85 & 34.46 & \textbf{35.31} & 35.10 & 34.87 & 34.21 & 33.53 & 32.71 & 32.02 & 33.48 & 33.63 & \textbf{36.76}\\
\hline
\hline
& \multicolumn{11}{c|}{\textbf{Summary Length (L)}} & \textbf{Best} & \textbf{Best} \\
\textbf{Newsroom} & 7 & 8 & 9 & 10 & 11 & 12 & 13 & 14 & 15 & 16 & \textbf{Avg.} & \textbf{Quality} & \textbf{Length}\\
\hdashline
\textbf{R-1 F$_1$} (\%) & 40.99 & 43.38 & 44.94 & 46.06 & 46.57 & \textbf{46.77} & 46.53 & 46.25 & 45.76 & 45.21 & 45.25 & 45.77 & 46.60\\
\textbf{R-2 F$_1$} (\%) & 19.15 & 20.99 & 22.11 & 23.02 & 23.47 & \textbf{23.59} & 23.38 & 23.15 & 22.79 & 22.33 & 22.40 & 22.58 & \textbf{23.85}\\
\textbf{R-L F$_1$} (\%) & 38.24 & 40.34 & 41.56 & 42.36 & \textbf{42.69} & 42.68 & 42.31 & 41.88 & 41.29 & 40.63 & 41.40 & 41.48 & \textbf{43.07}\\
\hline
\end{tabular}
\end{scriptsize}
\end{fontpbk}
\vspace{-0.05in}
\caption{Results on Gigaword and Newsroom datasets where the generator produces summaries of varying lengths.
}
\label{tab:results_varying_lengths}
\vspace{-0.15in}
\end{table*}

The system summaries are evaluated using both automatic metrics (ROUGE; Lin, 2004)\nocite{lin-2004-rouge} and human evaluation of information coverage, grammaticality and faithfulness to the original text.
We introduce a new dataset where a source sentence is associated with multiple summaries, and admissible ones are labelled by human annotators (\S\ref{sec:results}).
The dataset will serve as a useful testbed for future summarization research, where multiple reference summaries is key to improve the reliability of evaluation results~\cite{louis-nenkova-2013-automatically}.
This paper focuses on generating abstractive summaries from single source sentences. 
However, we expect the insights gained from this study to inform the design of future summarizers of different kinds.

\vspace{0.08in}
\noindent\textbf{Experimental Setup}\quad
Our generator is initialized with RoBERTa-\textsc{base}~\cite{liu2019roberta} due to its high performance on generation-related tasks.
We use Byte Pair Encoding~\cite{sennrich-etal-2016-neural} with a vocabulary of 50,265 tokens.
The model contains 12 Transformer blocks~\cite{NIPS2017_7181}, with a hidden size of 768 and 12 attention heads, for a total of 110M parameters.
We fine-tune the model on the train split of Gigaword and Newsroom, respectively, before applying it to the test sets. 
The model is fine-tuned for 20 epochs. 
Each epoch contains 24k / 1.5k batches and our batch size is 128. 
The model uses 10k / 1k warm-up steps, respectively, for Gigaword and Newsroom.
We use the AdamW~\cite{loshchilov2017decoupled} optimizer with an initial learning rate of 1e-4.
The momentum parameters are set to 0.9 and 0.999.
On a deep learning workstation equipped with 2x Titan RTX GPUs, our model takes 64 and 5.5 hours to fine-tune on Gigaword and Newsroom.
At test time, our beam size is $K$=20.
The model produces summaries ranging from $L$ = 7 to 16 tokens for a given source sentence.

Our selector for best overall quality is trained using 1.8M instances automatically constructed from the train split of Gigaword.
The set is balanced with an equal number of positive and negative instances.
226k instances are created with the type of Search and Replace, and 400k instances are created using each of the four remaining corruption types.
Our selector for best summary length is unsupervised and requires no training. 
The reward coefficient $r$ is set to 2.0 across all experiments.

\subsection{Experimental Results}
\label{sec:results}

\noindent\textbf{Automatic Evaluation}\quad\quad
In Table~\ref{tab:results_varying_lengths}, we present results on Gigaword and Newsroom test sets evaluated by ROUGE~\cite{lin-2004-rouge}.
We report R-1, R-2 and R-L $F_1$-scores that respectively measure the overlap of unigrams, bigrams, and longest common subsequences between system and reference summaries.
For each summarization instance, our generator produces multiple alternative summaries, ranging from $L$=7 to 16 tokens.
E.g., ``Daiwa Bank.'' corresponds to four tokens, `Dai', `wa', `Bank' plus an ending period.
Our \textsc{best-quality} and \textsc{best-length} selectors each identifies a single best summary from the set of alternative summaries for each summarization instance.

\begin{table*}[t]
\setlength{\tabcolsep}{6pt}
\renewcommand{\arraystretch}{1.2}
\centering
\begin{fontpbk}
\begin{scriptsize}
\begin{tabular}{|l||c|c|c|}
\hline
\textbf{Candidate Summary} & \textbf{Contains the main idea?} & \textbf{Is true-to-original?} & \textbf{Is grammatical?}\\
\hline
\hline
(1) Izetbegovic blasts Karadzic & $\square$ Yes \quad \done No & \done Yes \quad $\square$ No & \done Yes \quad $\square$ No\\
(2) Karadzic accused of swaying US Congress & \done Yes \quad $\square$ No & \done Yes \quad $\square$ No & \done Yes \quad $\square$ No \\
(3) Karadzic seeks to sway US Congress & \done Yes \quad $\square$ No & \done Yes \quad $\square$ No & \done Yes \quad $\square$ No \\
(4) Karadzic seeks to sway Congress & \done Yes \quad $\square$ No & \done Yes \quad $\square$ No & \done Yes \quad $\square$ No \\
(5) Karadzic misleading US Congress & $\square$ Yes \quad \done No & $\square$ Yes \quad \done No & $\square$ Yes \quad \done No \\
(6) Monday's international soccer scores & $\square$ Yes \quad \done No & $\square$ Yes \quad \done No & $\square$ Yes \quad \done No \\
\hline
\hline
\multicolumn{4}{|l|}{\textbf{Source Text:}
Bosnian President Alija Izetbegovic on Monday accused Bosnian Serb leader Radovan Karadzic of seeking}\\
\multicolumn{4}{|l|}{\,\,\quad\quad\quad\quad\quad\; to sway the US Congress against approving US troops to help enforce peace in the former Yugoslavia.}\\
\hline
\end{tabular}
\end{scriptsize}
\end{fontpbk}
\caption{
Example annotation interface.
A human annotator is instructed to read over the summaries before seeing the source text to effectively recognize any hallucinated content that is not found in the source text.
A native English speaker creates annotations for multiple instances, which are shared with all annotators to provide guidance.
}
\label{tab:annotation}
\vspace{-0.15in}
\end{table*}

We observe that the \textsc{best-length} selector has achieved the highest scores.
It performs better than using any single target length for all summaries.
Among summaries of different lengths, the highest R-2 $F_1$-scores are obtained when the target summary length is set to 11 and 12 tokens, respectively, for Gigaword and Newsroom.
This is close to the median length of reference summaries, which are 12 and 13 tokens for these datasets.
Our findings show that, 
the target summary length can make a non-negligible impact on automatic evaluation results.
It is best for system summaries to be long enough to include the most important information to achieve satisfying results.

In Table~\ref{tab:results_gigaword}, we report results on the Gigaword test split that contains 1,951 instances.
Our approach is compared against strong neural abstractive systems, including PEGASUS~\cite{zhang2020pegasus}, UniLM~\cite{NIPS2019_9464} and MASS~\cite{song2019mass}.
These systems draw on large-scale unsupervised pretraining to improve the quality of summaries, yielding some of the best reported results.
In comparison, our \textsc{best-length} selector either surpasses or performs comparably to these systems. 
The summaries selected by it achieve the highest R-2 $F_1$-score of 20.4\%.
We further choose the summary that yields the highest score for each instance, creating an oracle set of summaries, which yield a R-2 $F_1$-score of 33.4\%.
The results indicate that, with better summary selectors, there is a great potential that we can further boost summarization performance.

\begin{figure}[t]
\centering
\includegraphics[width=2.2in]{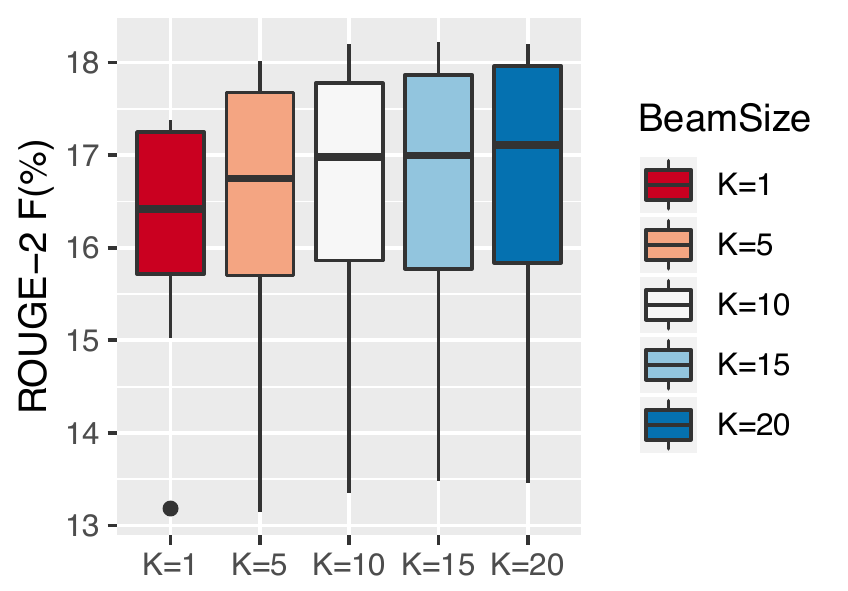}
\vspace{-0.05in}
\caption{Effectiveness of position-aware beam search (\S\ref{sec:decode}). 
A larger beam tends to give better results.}
\label{fig:beam}
\vspace{-0.15in}
\end{figure}

In Figure~\ref{fig:beam}, we investigate the effectiveness of our position-aware beam search (\S\ref{sec:decode}).
The beam size $K$ is set to $\{1, 5, 10, 15, 20\}$.
We report the average R-2 F$_1$-score across summaries of all lengths.
Results show that our position-aware beam search is effective at decoding summaries and works robustly across a range of beam sizes. 
A larger beam ($K$=20) tends to give better results.

\vspace{0.08in}
\noindent\textbf{Human Evaluation}\quad
We are interested in a holistic evaluation of the multiple alternative summaries produced by the generator.
To accomplish this, we develop a new dataset containing 500 summarization instances randomly sampled from the Gigaword test set.
Our generator produces 7 alternative summaries for each instance, which have varying lengths that range from $L$= 7 to 13 tokens.
We recruit human evaluators to judge the quality of each summary given its source text.\footnote{Our annotated dataset is available on Github at \url{https://github.com/ucfnlp/varying-length-summ}}

\begin{table}[t]
\centering
\setlength{\tabcolsep}{5pt}
\renewcommand{\arraystretch}{1.15}
\begin{fontpbk}
\begin{scriptsize}
\begin{tabular}{|l|cccc|}
\hline
& \textbf{Content} & \textbf{Truthful} & \textbf{Grammatical} & \textbf{Overall}\\
\hdashline
\textbf{Average} & 80.7 & 82.6 & 96.5 & 74.2 \\
\textbf{Best Length} & 82.8 & 86.0 & \textbf{97.4} & 77.8 \\
\textbf{Best Quality} & \textcolor{red}{\textbf{93.0}} & \textcolor{red}{\textbf{90.8}} & 97.0 & \textcolor{red}{\textbf{88.2}} \\
\hline
\end{tabular}
\end{scriptsize}
\end{fontpbk}
\caption{
Results of human assessment.
\textsc{best-quality} summaries have a higher likelihood of being admissible according to the criteria, suggesting the effectiveness of the method. 
}
\label{tab:results_human}
\vspace{-0.1in}
\end{table}

Our annotation interface is presented in Table~\ref{tab:annotation}.
A human annotator is instructed to read over all summaries before seeing the source text.
It allows him/her to effectively recognize any hallucinated content that is not found in the source text.
The annotator is asked to answer three yes-no questions. 
They include (a) has the summary successfully convey the main points of the source text? (b) does the summary preserve the meaning of the source? (c) is the summary grammatical?
A native speaker creates gold-standard annotations for multiple instances, they are shared with all annotators to provide guidance.
Our annotators are recruited using Appen (\url{appen.com}). It is a crowdsourcing platform similar to Amazon Mechanical Turk (\url{mturk.com}), but provides great quality control mechanisms to ensure high-quality work.

We recruit 5 annotators to judge the quality of each summary.
A summary is deemed \emph{admissible} under a criterion if the majority answer is yes.
We observe that, 74.2\% of summaries produced by our generator are admissible under all three criteria. 
The results suggest that our generator is able to produce multiple, equally valid summaries for a given source text.
We additionally examine the percentage of admissible summaries under each criterion, results are shown in Table~\ref{tab:results_human}.
Grammaticality has the best performance (96.5\%), followed by truthfulness (82.6\%) and content coverage (80.7\%). There appears to be room for improvement for the latter two aspects.
Moreover, the summaries chosen by our \textsc{best-quality} selector demonstrate a high admissible rate—93\%, 90.8\% and 97\%—respectively for the three criteria, suggesting the effectiveness of the selector.
Further, we observe a discrepancy between ROUGE and human judgments~\cite{fabbri2020summeval} as summaries yielding highest ROUGE scores are not always deemed admissible by human evaluators.
We hope this dataset provides a testbed for future summarizers to be judged on their ability to produce multiple summaries per instance rather than a single summary.

In Table~\ref{tab:example_order}, we show example system summaries and the order in which summary tokens are produced.
E.g., \{2,5\} indicate the two tokens ``Bo-J'' (Bank of Japan) are generated the 2nd and 5th place in the summary.
We find that our generator can effectively decide what content should be included in the summary given the limited space available, yielding summaries with varying degrees of details.
Important spans such as ``calls for calm'' tend to be generated first, less vital ones later. 
Our findings corroborate the hypothesis that a masked language model may enable generation in a flexible word order~\cite{liao-etal-2020-probabilistically}.
Further, we observe that the order in which tokens are generated is related to their dependencies (``call$\rightarrow$for''), which supports the findings of Clark et al.~\shortcite{clark-etal-2019-bert}.

\section{Conclusion}
\label{sec:conclusion}

We investigate a new approach to neural abstractive summarization that focuses on producing multiple summary hypotheses with varying lengths and diverse content.
Our selectors are designed to identify summaries that have the optimal length and the best overall quality.
The approach obtains state-of-the-art results on summarization benchmarks and opens up a potential new avenue for customizing summary selectors to suit users' needs.

Future work includes extending this research to long documents.
Our confidence-driven generator and the selectors could potentially be extended to operate on spans of text~\cite{joshi-etal-2020-spanbert} rather than individual tokens,
thus allowing for efficient generation of summary hypotheses that have varying degrees of details and identification of admissible summaries or summary segments.

\section*{Acknowledgements}
We are grateful to the reviewers for their insightful comments, which have helped us improve the paper.
This research was supported in part by the National Science Foundation grant IIS-1909603.

\bibliographystyle{acl_natbib}
\bibliography{fei,anthology,more}


\end{document}